\pdfoutput=1

\documentclass[11pt]{article}

\usepackage[]{acl}
\usepackage{xcolor,colortbl}
\usepackage{todonotes}
\usepackage{makecell}
\usepackage{times}
\usepackage{latexsym}
\usepackage{soul}
\usepackage[T1]{fontenc}

\usepackage[utf8]{inputenc}

\usepackage{amsmath}

\usepackage{microtype}
\usepackage{graphicx}
\usepackage{tabularx}
\usepackage{makecell}
\usepackage{multirow}
\usepackage{subcaption}
\usepackage{enumitem}

\title{Consultation Checklists: Standardising the Human Evaluation\\of Medical Note Generation}

\author{Aleksandar Savkov\textsuperscript{1}, Francesco Moramarco\textsuperscript{1,2}, Alex Papadopoulos Korfiatis\textsuperscript{1}, \\
\bf{Mark Perera},
\bf{Anya Belz\textsuperscript{2,3}},
\bf{Ehud Reiter\textsuperscript{2}}
\\
\textsuperscript{1}Babylon
\textsuperscript{2}University of Aberdeen
\textsuperscript{3}ADAPT Research Centre, Dublin City University
\\
\textsuperscript{1} \texttt{\{sasho.savkov, francesco.moramarco, alex.papadopoulos\}}
\\ 
\texttt{@babylonhealth.co.uk}
\\
\textsuperscript{2} \texttt{\{r01fm20, ehud.reiter, anya.belz\}@abdn.ac.uk}}

\begin{document}
\maketitle

\begin{abstract}
Evaluating automatically generated text is generally hard due to the inherently subjective nature of many aspects of the output quality. This difficulty is compounded in automatic consultation note generation by differing opinions between medical experts both about which patient statements should be included in generated notes and about their respective importance in arriving at a diagnosis. Previous real-world evaluations of note-generation systems saw substantial disagreement between expert evaluators.
In this paper we propose a protocol that aims to increase objectivity by grounding evaluations in Consultation Checklists, which are created in a preliminary step and then used as a common point of reference during quality assessment. 
We observed good levels of inter-annotator agreement in a first evaluation study using the protocol; further, using Consultation Checklists produced in the study as reference for automatic metrics such as ROUGE or BERTScore improves their correlation with human judgements compared to using the original human note.

\end{abstract}

\renewcommand{\cellalign}{l}

\begin{table}[t]
\small
    \setlength{\tabcolsep}{5pt} 
    \def\arraystretch{1.5}

    \centering
    \begin{tabular}{p{4cm}|p{3.3cm}}
        \multicolumn{1}{c|}{\cellcolor{blue!25}\textbf{Transcript}} & \multicolumn{1}{c}{\cellcolor{blue!25}\textbf{Note}} \\\hline
                  \cellcolor{blue!5}Clinician: Hello there, it's Dr Smith, and how can I help you this afternoon? & \multirow{8}{*}{\makecell{
3/7 hx developed headache.\\
Constant, severity 8/10, \\
dull ache with associated \\
sharp pain, gradual onset. \\
Progressively worsening. \\
Has tried ibuprofen with \\ 
limited relief.\\
Feels nauseous, no vomit.\\
No neck pain/stiffness.\\
No speech disturbances.\\
No arm or leg weakness.\\
No head injury. No fevers.\\ 
No rashes. \\
PMH: Nil. \\
DH: Nil. NKDA \\
FH: mother and sister - \\
migraines\\
SH: lives with \\
housemates, works in IT \\
Socially smoke/EtOH.}}\\\cline{1-1}
         
Patient: Hi there. Well, I have this like really crazy headache that's been going on for days.\\\cline{1-1}
\cellcolor{blue!5}Clinician: Ohh dear, OK. When, when did it exactly start, this headache?\\\cline{1-1}
Patient: Eh, around three days ago, maybe.\\\cline{1-1}
\cellcolor{blue!5}Clinician: Three days ago, OK. And whereabout in your head, is this pain?\\\cline{1-1}
Patient: Um, it kind of feels all over my head, but mainly around my right eye. [...]\\
    \end{tabular}
    \caption{Abridged version of a mock transcript and human-written note from \citet{korfiatis2022primock57}.}
    \label{tab:transcript_note}
\end{table}

\section{Introduction}
\label{sec:introduction}
While Electronic Health Record systems are a necessity in modern healthcare, they are burdening primary care clinicians with significant clerical work that distracts them from patient care and increases their dissatisfaction and burnout rates \citep{arndt_tethered_2017}.
Since a significant part of the required documentation involves note writing, there has been a mounting interest in assisting clinicians by automatically generating consultation notes \cite{finley_automated_2018,enarvi_generating_2020, molenaar_medical_2020,knoll2022user}.

A common approach involves passing the recording of the consultation through a speech-to-text system, then using a sequence-to-sequence model trained on parallel transcript and note datasets to automatically generate the note \cite{krishna2020generating, joshi2020dr, zhang2021leveraging, moramarco2022human}. An example of a transcript and associated consultation note, taken from the mock consultation dataset released by \citet{korfiatis2022primock57}, can be seen in Figure \ref{tab:transcript_note}.


Evaluating the output of such systems is challenging \cite{gehrmann2022repairing}, as it is often the case in Natural Language Generation (NLG). Widely used automatic metrics, such as ROUGE \cite{lin2004rouge} and BLEU \cite{papineni2002bleu}, often fail to capture relevant aspects of generated text \cite{reiter2009investigation}, and human evaluation, the best practice in NLG, is not only expensive and hard to reproduce \cite{belz2021reprogen} but also highly subjective \cite{howcroft2020twenty, VANDERLEE2021101151, gehrmann2022repairing}. Even in the field of Note Generation where evaluators tend to be medical experts rather than crowd-sourced workers, inter-annotator agreement is low, as there is no explicit ground truth and the annotators have differing opinions on the importance of each patient statement and whether it should be included in a consultation note \mbox{\cite{moramarco2022human}}.


In this work, we propose an evaluation protocol that uses Consultation Checklists (CC), itemised reference of all facts discussed during doctor-patient consultations. We report good agreement between clinicians building CCs from the same consultation, which indicates good consistency of the reference creation process.
Since Consultation Checklists act as an approximation of the ground truth, they reduce evaluator subjectivity, which is reflected in the high inter-annotator agreement observed in our first study. We also show that correlation with human judgements increases when using CCs instead of the original clinician note as the reference for automatic evaluation metrics. 

\section{Related Work}
\label{sec:related_work}
There are a number of different approaches to quantitative human evaluation in NLG.

\begin{description}[leftmargin=*]
    \itemsep0em
    \item[Rating or Likert scales] work well with few criteria, but lack explanatory power and fail to capture text quality \citep{hastie2014comparative}. Adding more criteria partly resolves this, but at the cost of the evaluation task becoming more difficult and subjective \citep{VANDERLEE2021101151}. For example, \citet{MOEN201625} use a 30 item rating scale
and report that subjects found it too difficult to use.

    \item [Ranking methods,] where evaluators are asked to rank the output of text generation systems along a specified criterion, are an alternative to rating scales. Some studies have shown ranking to be more reliable and consistent \citep{VANDERLEE2021101151}; however, ranking methods do not scale well when comparing multiple models.

    \item[Extrinsic measures,] such as measuring post-edit time of generated text \citep{moramarco2022human} provide a better estimate of how useful the generated text may be to the final user, but are often expensive and subjective \cite{lai2022exploration}.
\end{description}

A common shortcoming among all methods described above is that none of them provide granular insights into the text generation systems' errors and how to address them. This is particularly important when evaluating automatically generated medical notes, where the factual accuracy and completeness of the generated note are critical, as well as identifying the situations where the system fails.

As with all summarisation tasks, Note Generation has an element of content selection that is highly subjective. For this reason, evaluating system-generated notes against a single reference summary would penalise those notes that diverge from the reference in their content selection. 
One way to address this is by using multiple reference summaries\footnote{Multiple reference summaries are also used in some automatic evaluation metrics, such as ROUGE \citep{lin2004rouge}.}, as for example in the Pyramid evaluation protocol \citep{nenkova2004evaluating}. In a similar way to our proposed protocol, it splits the evaluation into two independent steps: extracting Summarization Content Units (SCUs) from multiple references, then using these SCUs to evaluate generated text.




Another way to address the subjectivity of using single references comes from \emph{reference-less} approaches that compare the generated text against the source text directly rather than against reference summaries. For example, \citet{narayan2019highres} once more split the evaluation in two steps: highlighting annotations in the source document, then comparing each generated summary to these annotations. In the domain of note generation, however, the format of the source documents (consultation transcripts) and the generated summaries (consultation notes) is different enough (see Table \ref{tab:transcript_note}) that a highlighting-based approach would not work.

\begin{figure*}[!ht]
    \centering
    \includegraphics[width=.9\textwidth]{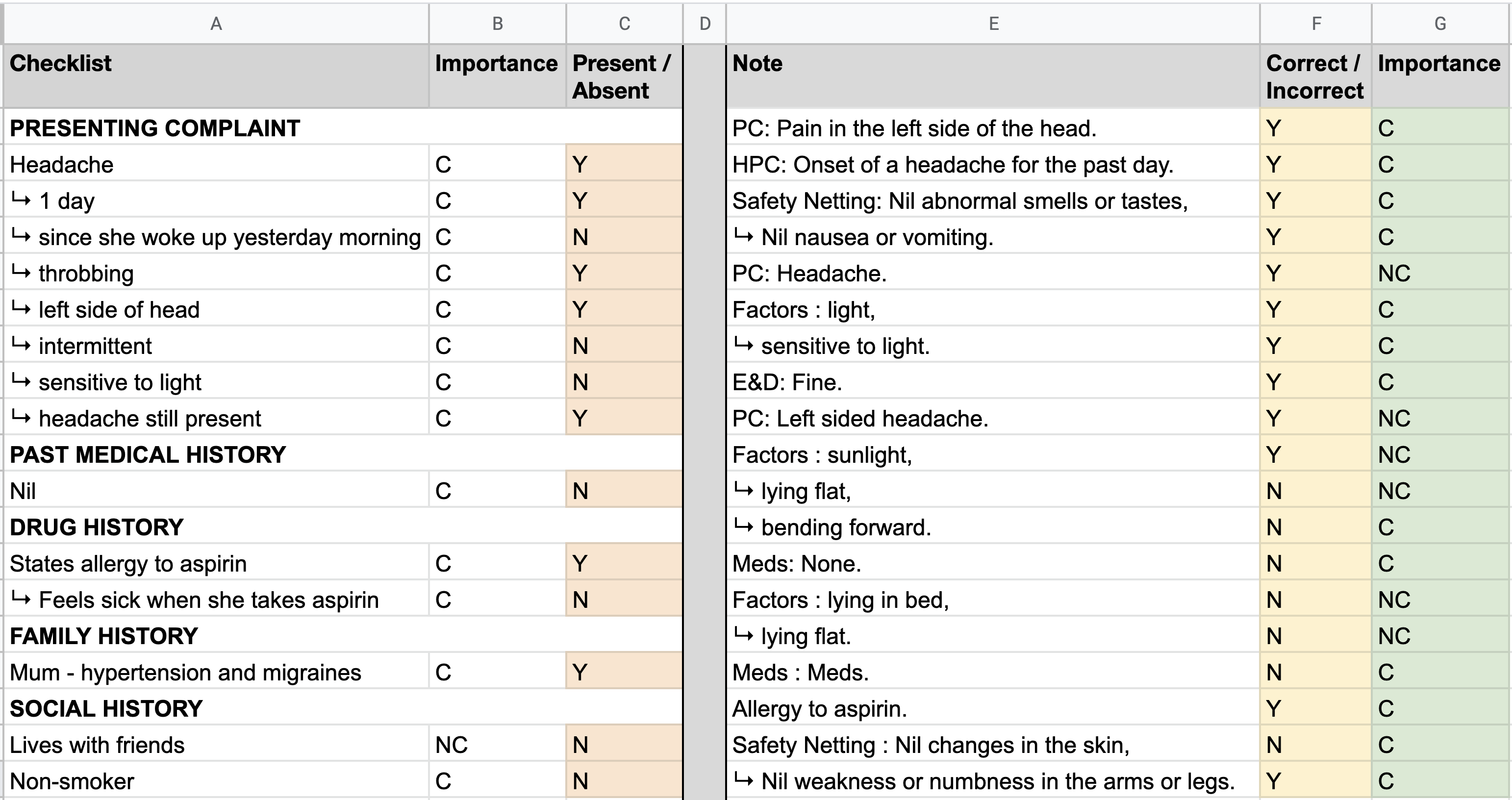}
    \caption{Columns A and B are the abridged version of a Consultation Checklist. Column E is the automatically-itemised system generated note. Columns C, F, and G are filled in by the evaluating clinician.}
    \label{fig:note-evaluation}
\end{figure*}

In the medical domain \citet{moramarco2022human} give evaluators the consultation audio recording instead of a reference, and ask them to identify the missing and incorrect items in a generated note, providing the required insight into how the generated text is wrong.
However, even though the evaluators are experts (medical practitioners), agreement between them is very low. While this could be improved by better evaluator training, we believe that the evaluation task itself inherently inhibits agreement. There is no standard way of recognising or mapping facts from the audio recording of the consultation, which is the `ground truth', to the consultation note -- generated or otherwise. This makes it very hard to align multiple evaluators and get consistent results.


\section{Proposed Protocol}
\label{sec:method}



\begin{figure}[t]
    \centering
    \includegraphics[width=.38\textwidth]{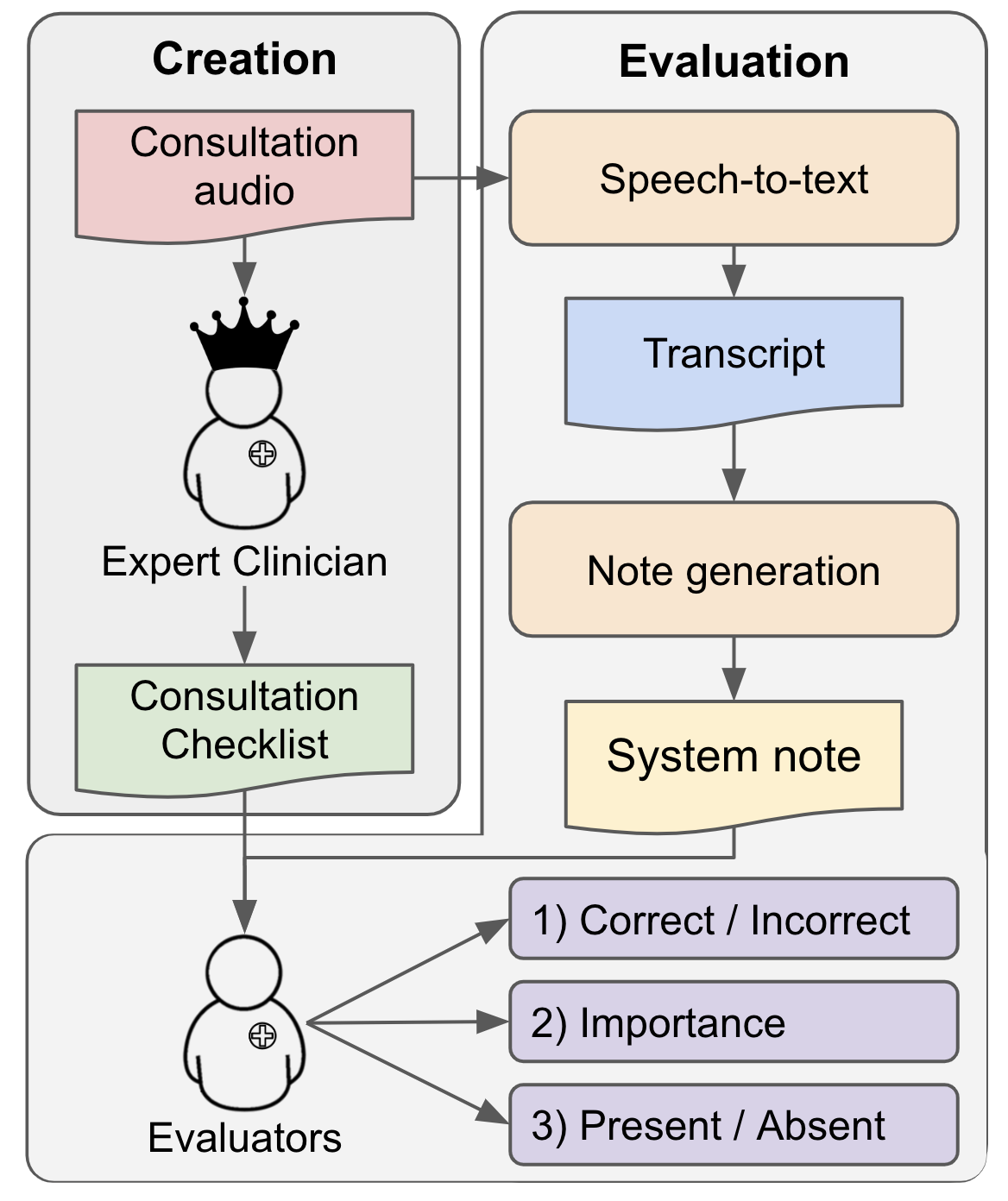}
    \caption{Diagram of the Consultation Checklists evaluation protocol, including the creation stage and the evaluation stage.}
    \label{fig:diagram}
\end{figure}

We propose Consultation Checklists --- a protocol using an expert-crafted ground truth approximation to evaluate the quality of system-generated medical notes with human raters.
The evaluation protocol consists of a reference creation step followed by a notes evaluation step (see \mbox{Figure \ref{fig:diagram}}).

\subsection{Creation}
Given a dataset of consultation audio recordings, one expert clinician is asked to listen to the audio and produce a Consultation Checklist: a structured list of \textbf{all the facts} discussed in the consultation. Including both relevant and irrelevant content in the Consultation Checklist is an important feature of the protocol as it eliminates the subjectivity of the content selection characteristic of other human-made references as discussed in Section \ref{sec:related_work}.
The list items are organised in sections for clarity and split into sub-lists to allow for more granularity (e.g. `headache for 1 day' may be item `Headache' with sub-item `1 day'); see Columns A and B in \mbox{Figure \ref{fig:note-evaluation}.}
Following \citet{moramarco2022human}, each item is marked for clinical importance as follows:


\begin{description}[leftmargin=*]
    \itemsep0em
    \item [Critical:] Items medico-legally required to document the diagnosis and treatment decisions whose absence or incorrectness may lead to wrong diagnosis and treatment later on, e.g. the symptom `cough' in a suspected chest infection consultation. This is the key information a note needs to capture correctly in order to not mislead clinicians.

    \item [Non-critical:] Items that should be documented in a complete note but whose absence will not affect future treatment or diagnosis, e.g. `who the patient lives with' in a consultation about chest infection.
    
    \item [Irrelevant:] Medically irrelevant information covered in the consultation, e.g. the pet of a patient with a suspected chest infection just died. Including such information in the Consultation Checklist allows for a fair evaluation of the less relevant parts of the generated notes. 

\end{description}

\noindent Once the Consultation Checklists are created, they can be stored and reused in any future evaluation, thereby making the evaluation cost more scalable.

\subsection{Evaluation}
The created Consultation Checklists are then used to evaluate one or more system-generated consultation notes by one or more clinicians (or raters). These clinicians are not required to listen to the consultation recording but to only rely on the Consultation Checklist as a common ground when evaluating the notes. 
Each note is automatically split into sentences, then each sentence is split on punctuation and conjunctions.
The first item in each sentence is considered top level and all others are nested in a sub-list (column E in Figure \ref{fig:note-evaluation}). We instruct evaluators to read the full sentence before marking each item as they may be meaningless in isolation.
Once familiar with the Consultation Checklist, clinicians are asked to carry out the following sequence of tasks (see Figure \ref{fig:diagram}):

\begin{enumerate} [leftmargin=*]
    \itemsep0em
    \item Mark each item in the note as \emph{correct} or \emph{incorrect} using the Consultation Checklist as a common reference (column F). Since there is no explicit mapping, the clinicians need to scan the Consultation Checklist to try and find supporting facts for each item in the note. For example, in Figure \ref{fig:note-evaluation} item 2 in the note (`HPC\footnote{History of Presenting Complaint}: Onset of a headache for the past day') is validated by the first two items in the Consultation Checklist (`Headache' and '1 day').

    \item Mark each item in the generated note as critical, non-critical, or irrelevant (column G). Even though some of these values could be inferred from the importance of the Consultation Checklist items, asking the evaluating clinicians to fill them in covers two common edge cases: (i) when one item in the note is fact-checked against multiple items in the Consultation Checklist, and (ii) when the item is incorrect. In the case of incorrect items, the importance is established by the effect the presence of the item may have to the clinician using the generated note.
    
    \item Mark each item in the Consultation Checklist as present or absent in the generated note (column C) where `present' means that the item is fully reported in the generated note.

\end{enumerate}

\noindent We define \textit{Precision} and \textit{Recall} in the context of Consultation Checklists as:

\begin{equation}
    Precision = \frac{|\ correct\ items\ |}{|\ generated\ items\ |}
\label{eq:precision}
\end{equation}

\begin{equation}
Recall = \frac{|\ present\ items\ |}{|\ checklist\ items\ |}
\label{eq:recall}
\end{equation}


\noindent Both metrics can also be computed for critical items only using the importance level assigned to each item (see Figure \ref{fig:note-evaluation}, column B). 

The items in each Consultation Checklist are similar to the `Summarization Content Units' described by \citet{nenkova2004evaluating} and can be re-used to evaluate any number of generated notes. However, as our method does not use human notes as a reference, the items are extracted from the consultation audio recording rather than from multiple reference summaries. As the evaluation stage involves no writing and only a limited amount of interpretation, the set of clinicians carrying it out could be of lower skill in contrast to the Pyramid approach where the same amount of skill is required for both stages.



\section{Checklist Creation Pilot}
\label{sec:checklist-creation}


Based on our initial assumption that a Consultation Checklist should capture the salient points of a consultation in an itemised format, we ran a pilot study with 2 clinicians expert in AI annotation, A \& B. The goal of the study was to define best practices for creating Consultation Checklists and to evaluate the consistency of their creation between clinicians.


To investigate the agreement on Checklist creation, we asked the two clinicians to produce Consultation Checklists for the same 10 mock consultations taken from \citet{korfiatis2022primock57} (see Figure \ref{fig:checklist-alignment} in the Appendix for an example).
In order to quantify the alignment between them, two of the authors checked whether the information of each item in Clinician A's version was present in Clinician B's version, and vice-versa.

Based on this analysis, we define the fact coverage for each Consultation Checklist as the number of matching facts divided by the total number of facts. We found that 93.7\% of the items in Clinician A's Consultation Checklists were also present in Clinician B's, and 78.3\% the other way around. This may indicate that the two clinicians agree on the basic facts to include but Clinician B tended to add more details.
Table \ref{tab:checklist-alignment} shows the values for both annotators, the average, and the agreement computed with Krippendorff's Alpha on the binary values for each statement (present or absent).

\begin{table*}[h!]
    \centering
    \begin{tabular}{l|l|l}
        \textbf{Annotator} & \textbf{Checklist A (critical)} & \textbf{Checklist B (critical)}\\\hline
        \textbf{Ann1} & 94.7\% (95.7\%) & 79.8\% (79.1\%)\\
        \textbf{Ann2} & 92.8\% (94.3\%) & 76.9\% (78\%)\\\hline
        \textbf{Avg}  & 93.7\% (95\%)   & 78.3\% (78.6\%)\\\hline
        \textbf{Agreement} & 0.624 & 0.77\\
    \end{tabular}
    \caption{Results of the alignment between Checklists.}
    \label{tab:checklist-alignment}
\end{table*}

As part of the pilot, we refined the following process for creating Consultation Checklists:
\begin{enumerate}[leftmargin=*]
    \itemsep0em
    \item Listen to the consultation audio and take notes on every patient statement.
    \item Format the notes into an itemised list, splitting longer items to a more atomic level in sub-lists and categorising them using sub-headers (e.g. Presenting History, Past Medical History, Social History, etc.). More examples of sub-headers can be seen in Figure \ref{fig:checklist-alignment} (Appendix \ref{sec:appendix}).
    \item Read through two system-generated notes to sanity-check the Checklist with regards to item granularity and coverage.
    \item Mark each item in the Checklist as critical, non-critical or irrelevant, as defined in Section \ref{sec:method}.
    \item Re-listen to the consultation audio to ensure no important points have been missed.
\end{enumerate}

\noindent The pilot study also highlighted the cognitive effort of producing Consultation Checklists. On average, a Consultation Checklist for a 10 minutes consultation contains 56 items (excluding sub-headings) and it takes the clinician around 1 hour to complete. Also, the clinicians found that they would be able to produce 4 Consultation Checklists in a row before requiring a break, and that the first one would be the quickest to make (in as little as 30 minutes) but the following ones would require progressively more time.



\section{Consultation Notes Evaluation}
\label{sec:evaluation-experiments}
To test the utility of our protocol, we used a single expert clinician to create 20 Consultation Checklists from real-life patient consultations.
We then generated History and Examination notes for each of the consultations using a BART \cite{lewis2020bart} encoder-decoder transformer model\footnote{\url{https://huggingface.co/facebook/bart-large-cnn}}. The model was pre-trained on the CNN/Dailymail dataset \citep{hermann2015teaching}, and fine-tuned on a proprietary dataset of 10,000 (speech-to-text generated) consultation transcripts and human-written History and Examination notes.

Finally, we hired three clinicians (two women and one man from ethnically diverse backgrounds) to evaluate the 20 generated notes by following the evaluation process defined in Section \ref{sec:method}. The training for this task involved: (i) task instructions (see Appendix); (ii) two evaluation practice tasks; (iii) an alignment session between the three clinicians, where they investigated all cases of disagreement on the practice tasks and came to a joint decision.

\begin{table}[!ht]
    \centering
    \begin{tabular}{l|l|l|l}
        & \thead{\textbf{Present /}  \\ \textbf{Absent}} & \thead{\textbf{Correct /} \\ \textbf{Incorrect}} & \thead{\textbf{Imp.}} \\\hline
        \textbf{\# data points}  & 2258 & 904 & 904 \\\hline
        \textbf{Eval 1 - Eval 2}  & 0.733 & 0.690 & 0.521 \\
        \textbf{Eval 1 - Eval 3}  & 0.729 & 0.627 & 0.387 \\
        \textbf{Eval 2 - Eval 3} & 0.754 & 0.701 & 0.697 \\ \hline
        \textbf{3-way Agreement} & 0.739 & 0.672 & 0.522 \\ \hline
    \end{tabular}
    \caption{Krippendorf's Alpha inter-annotator agreement scores for Present/Absent, Correct/Incorrect and Importance (Imp.).}
    \label{tab:agreement}
\end{table}

\begin{table}[!ht]
    \centering
    \begin{tabular}{l|l|l|l}
        & \thead{\textbf{CC} \\ \textbf{(pairwise)}} & \thead{\textbf{CC} \\ \textbf{(count)}} & \thead{\textbf{No CC} \\ \textbf{(count)}}\\\hline
        \thead{\textbf{Pre / Abs}} & 0.739 & 0.969 & 0.374* \\
        \thead{\textbf{Cor / Inc}} & 0.672 & 0.726 & 0.541* \\
    \end{tabular}
    \caption{Krippendorff's Alpha scores for Present/Absent and Correct/Incorrect values for an evaluation using the Consultation Checklist as common ground and one using the consultation recording. Asterisk(*) denotes scores reported in \citet{moramarco2022human}.}
    \label{tab:agreement-checklist}
\end{table}

In addition, we wanted to quantify how often evaluators misjudge the generated note because of information that is omitted or misrepresented in Consultation Checklists. In order to do this, we asked each evaluator to listen to the audio of the consultation and review whether a generated note item was correct or incorrect.


\begin{table*}[!ht]
    \centering
    \begin{tabular}{l|l|l|l|l}
        & \multicolumn{2}{c|}{\textbf{Spearman correlation coefficient}} & \multicolumn{2}{c}{\textbf{Pearson correlation coefficient}} \\
        \textbf{Metric} & \textbf{Ref: Human note} & \textbf{Ref: Checklist} & \textbf{Ref: Human note} & \textbf{Ref: Checklist}\\ \hline
        Rouge1 Fscore  & 0.493 & 0.553 & 0.509 & 0.553\\
        Rouge2 Fscore  & 0.376 & 0.570 & 0.445 & 0.545\\
        Rouge3 Fscore  & 0.348 & 0.424 & 0.408 & 0.442\\
        RougeL Fscore  & 0.431 & 0.636 & 0.439 & 0.577\\
        BERTScore      & 0.375 & 0.58 &  0.414 & 0.563\\
        Levenshtein Distance\textdagger & 0.003 & 0.284 & 0.064 & 0.224\\
    \end{tabular}
    \caption{Spearman and Pearson correlation coefficients against the human judgements (an average of precision and recall) from the study. \textdagger\ denotes lack of statistical significance ($p \ge 0.05$).}
    \label{tab:correlation}
\end{table*}

\section{Results \& Discussion}
\label{sec:results}


\subsection{Inter-Annotator Agreement}
One of the advantages of the Consultation Checklists protocol is that it allows us to compute inter-annotator agreement at a item level, as opposed to just error counts as done in \citet{moramarco2022human}. Table \ref{tab:agreement} shows Krippendorff's Alpha\footnote{\url{https://pypi.org/project/krippendorff/}} \cite{krippendorff2018content} scores for the raw pairwise values of \textit{Present / Absent}, \textit{Correct / Incorrect}, and \textit{Importance}. We use nominal agreement for the first two, which have binary values and ordinal agreement for Importance by converting \textit{irrelevant}, \textit{non-critical}, and \textit{critical} to integers.

In order to compare agreement to \citeauthor{moramarco2022human}'s reported results, we also compute interval agreement on error counts. While the results are not directly comparable since the dataset, sample size and annotator count are different, the agreement using the Consultation Checklists protocol is much higher (Table \ref{tab:agreement-checklist}). It is also a considerable increase over the average NLG human evaluation agreement of 0.3 to 0.5 reported by \citet{VANDERLEE2021101151}.

\subsection{Accuracy Trade-off}
As mentioned in Section \ref{sec:method}, when generating Consultation Checklists, the goal is to capture as much of the consultation as possible. However, it is difficult to capture all points while keeping the Consultation Checklist concise, and some nuances which might be needed to faithfully assess the generated notes could be missed.

This trade-off between evaluation accuracy and standardisation is a limitation of our approach that we quantified by checking how often evaluators changed their Correct / Incorrect answers after listening to the consultation audio (Table \ref{tab:changes}). On average, this only happened for a small number of generated note items (3.91\%). Most changes were from Incorrect to Correct, which highlights the importance of making sure the Consultation Checklists are a thorough representation of the consultation.
For example, consider a checklist that includes the item ``was feeling cold'' but omits the extra information of ``had to wear more clothes than usual''. In this case, a generated note item referring to this extra information would be marked as Incorrect based on the Consultation Checklist, but as Correct based on the audio.

\begin{table}[h!]
    \centering
    \begin{tabular}{l|l|l|l}
        \textbf{Evaluator} & \textbf{\# Changes} & \thead{\textbf{Correct ->} \\ \textbf{Incorrect}} & \thead{\textbf{Incorrect ->} \\ \textbf{Correct}}\\ \hline
        Eval 1 & 44 (4.87\%)  & 9 & 35\\
        Eval 2 & 39 (4.31\%) & 4 & 35\\
        Eval 3 & 23 (2.54\%) & 3 & 20\\ \hline
        Avg & 35 (3.91\%) & 5.33 & 30 \\
    \end{tabular}
    \caption{Changes in Correct / Incorrect values after listening to the consultation recording.}
    \label{tab:changes}
\end{table}

\subsection{Time Efficiency}

It took clinicians a self-reported 45 minutes on average to complete each evaluation task: 5 minutes to understand the Consultation Checklist; 15 minutes for each of the two notes to evaluate correct vs. incorrect and present vs. absent items, including item importance; and 10 minutes to listen to the consultation audio and modify any answers. For context, \citet{moramarco2022human} report that their evaluators need 1 hour to listen to a consultation audio recording and evaluate 5 consultation notes.


\subsection{Consultation Checklists for automatic evaluation}

Since Consultation Checklists aim to be a standardised reference, we expected that their textual representation\footnote{We get the textual representation of a Consultation Checklist by concatenating all items in a single string.} can also be used as a more objective reference than a single clinician's consultation note for automatic metrics.
To test this, we computed the scores for a few common NLG metrics: ROUGE \cite{lin2004rouge}, BERTScore \cite{zhang2019bertscore} and Levenshtein distance \cite{levenshtein1966binary} using either clinician-written notes or Consultation Checklists as the reference text.
Table \ref{tab:correlation} 
reports both Spearman and Pearson correlation coefficients against our human judgements (an average of precision and recall). For all three metrics, using Consultation Checklists as references increases the correlation with human judgements, with BERTScore showing the highest gain at 20.5\% Spearman's correlation increase.
\section{Conclusion}
\label{sec:conclusion}


In this work we proposed a novel reference data structure called Consultation Checklists (CC), and a protocol that uses it for evaluating automatically generated consultation notes. Our experiments show good inter-annotator agreement levels when parallel CCs are created from the same set of clinical consultations. We also report good agreement when the Consultation Checklist protocol is used by different clinicians to evaluate the same consultations.
Finally, we showed that expertly-crafted Consultation Checklists are better than human-written notes when used as references for automatic evaluation.

While we have tested our protocol only on note generation for primary care consultations, we postulate that consultation checklists would apply to a number of medical domains, including secondary care and 

\section{Ethical Considerations}
We considered the ethical implications of this work and found no concerns. The study participants are senior clinicians with at least 5-10 years of experience consulting. They are paid £70 an hour, have agreed to work for a maximum of 8 hours per week, and able to withdraw at any time. The consultations they are asked to evaluate are real doctor-patient interactions. These are stored securely following GDPR practices and the patients have consented for their data to be used for research purposes.

Finally, while we hope it can be generalised and applied to other domains, medical and otherwise, the evaluation protocol we propose in this paper has only been tested in the domain of primary care UK consultations.






\bibliography{anthology,custom}
\bibliographystyle{acl_natbib}

\appendix
\onecolumn
\section{Appendix}
\label{sec:appendix}

\subsection{Evaluating Clinicians instructions}
\label{app:eval-instructions}


\begin{enumerate}
    \item Read through the checklist (on the left side of the spreadsheet) and familiarise yourself with the consultation.

    \item Based on the information in the checklist, go through the generated note (on the right side of the spreadsheet) and mark each statement of the note as \textbf{correct} (y) / \textbf{incorrect} (n). [Column G]

    \item Mark each statement in the note as either critical, non critical or irrelevant: [Column I]
\begin{enumerate}
    \item \textbf{Critical}: The statement is of critical medical importance. If it’s a correct statement, the note would not be medically complete without the statement present; if it’s an incorrect statement, its presence in a consultation note would be a medical risk (for example, could lead to a different diagnosis).
    \item \textbf{Non-critical}: The statement is of medical value, but its presence or absence in the note is not critical medically. For example, some doctors might include a non-critical statement in their note but other doctors could skip it.
    \item \textbf{Irrelevant}: The statement is irrelevant; if correct, most doctors would not include it in the note (for example, “Patient reports they prefer to wear green clothes”). If incorrect, its presence in the note is inconsequential.
\end{enumerate}
    \item Go through each statement in the checklist and mark it as either present or absent in the generated note. [Column C]

    \item Repeat this for each generated note

    \item Now, listen to the actual consultation recording. Take notes if you need to, especially if something you hear is different from what you understood through reading the consultation checklist.

    \item Finally, fill the ``Correct / Incorrect (after audio)'' field, essentially amending your earlier answers after listening to the consultation audio. This will allow us to evaluate how much information and context is lost by using the consultation checklist instead of the actual recording. [Columns D,I]

\end{enumerate}

\newpage
\subsubsection{Checklists comparison}



\begin{figure*}[h!]
    \centering
    \includegraphics[width=1\textwidth]{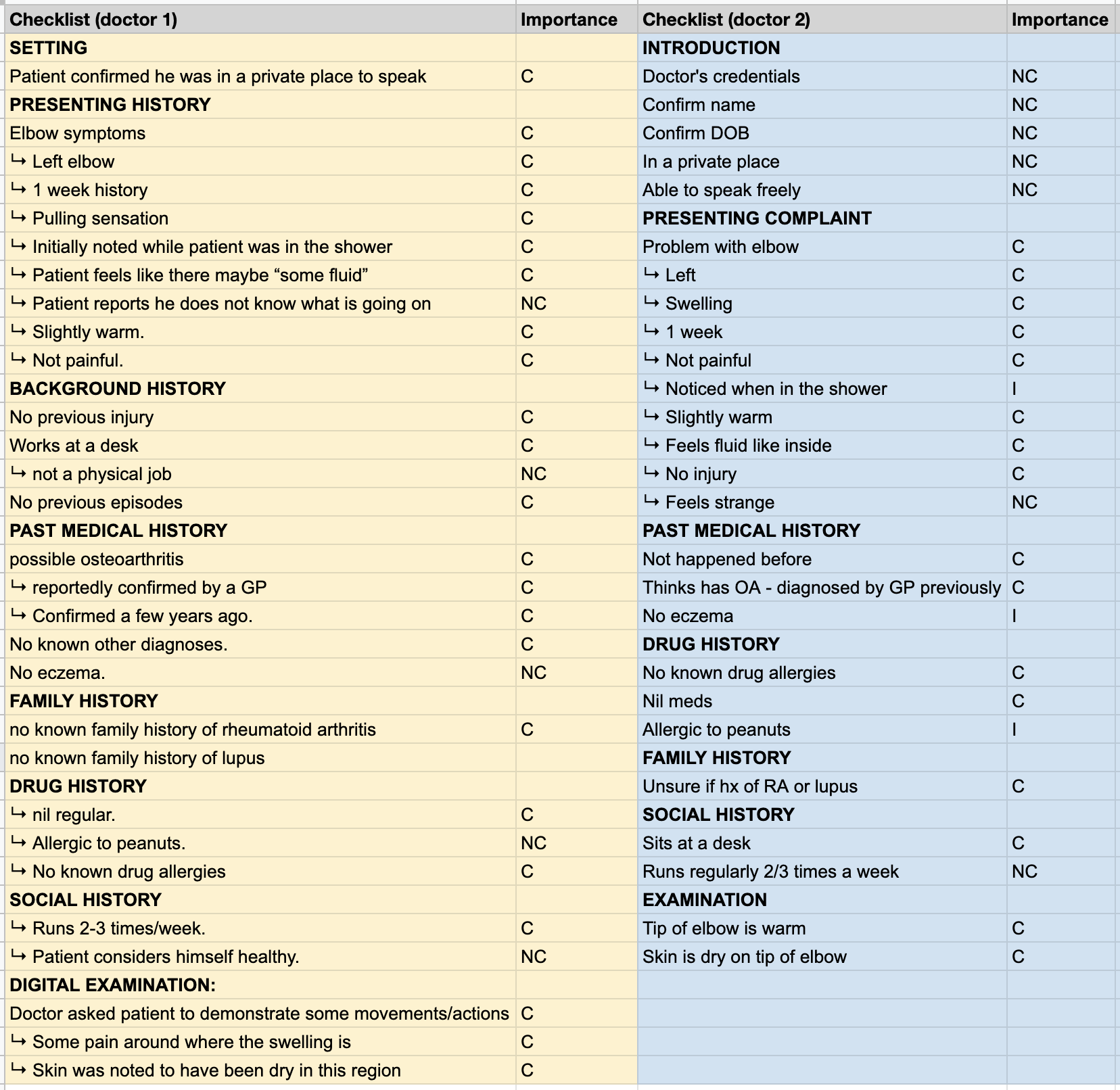}
    \caption{Example of two checklists for the same mock consultation.}
    \label{fig:checklist-alignment}
\end{figure*}

\newpage

\end{document}